\documentclass[10pt,twocolumn,letterpaper]{article}

\usepackage{wacv}
\usepackage{times}
\usepackage{epsfig}
\usepackage{graphicx}
\usepackage{amsmath}
\usepackage{amssymb}
\usepackage{subfigure}
\usepackage[ruled,lined,linesnumbered]{algorithm2e}

\usepackage[english]{babel}
\usepackage{balance}

\def\wacvPaperID{212} % Enter the WACV Paper ID here

\wacvfinalcopy % *** Uncomment this line for the final submission

\ifwacvfinal
\usepackage[breaklinks=true,bookmarks=false]{hyperref}
\else
\usepackage[pagebackref=true,breaklinks=true,colorlinks,bookmarks=false]{hyperref}
\fi

\begin{document}
%%%%%%%%% TITLE
\title{Covariance-free Partial Least Squares: \\An Incremental Dimensionality Reduction Method}

\author{Artur Jordao, Maiko Lie, Victor Hugo Cunha de Melo and William Robson Schwartz\\
Smart Sense Laboratory, Computer Science Department\\
Federal University of Minas Gerais, Brazil\\
{Email: \{arturjordao, maikolie, victorhcmelo, william\}@dcc.ufmg.br}
}

\maketitle
%\thispagestyle{empty}

%%%%%%%%% ABSTRACT
\begin{abstract}
Dimensionality reduction plays an important role in computer vision problems since it reduces computational cost and is often capable of yielding more discriminative data representation. In this context, Partial Least Squares (PLS) has presented notable results in tasks such as image classification and neural network optimization. However, PLS is infeasible on large datasets, such as ImageNet, because it requires all the data to be in memory in advance, which is often impractical due to hardware limitations. Additionally, this requirement prevents us from employing PLS on streaming applications where the data are being continuously generated. Motivated by this, we propose a novel incremental PLS, named \emph{Covariance-free Incremental Partial Least Squares} (CIPLS), which learns a low-dimensional representation of the data using a single sample at a time. In contrast to other state-of-the-art approaches, instead of adopting a partially-discriminative or SGD-based model, we extend Nonlinear Iterative Partial Least Squares (NIPALS) --- the standard algorithm used to compute PLS --- for incremental processing.  Among the advantages of this approach are the preservation of discriminative information across all components, the possibility of employing its score matrices for feature selection, and its computational efficiency. We validate CIPLS on face verification and image classification tasks, where it outperforms several other incremental dimensionality reduction techniques. In the context of feature selection, CIPLS achieves comparable results when compared to state-of-the-art techniques.
\end{abstract}
%\keywords{Incremental Dimensionality Reduction \and Partial Least Squares}
\section{Introduction}\label{sec:introduction} Dimensionality reduction is widely used in computer vision applications from image classification~\cite{Hasegawa:2016}~\cite{Alakkar:2019} to {detection of adversarial images}~\cite{Hendrycks:2017}. The idea behind this technique is to estimate a transformation matrix that projects the high-dimensional feature space onto a low-dimensional latent space~\cite{Martinez:2001}\cite{Geladi:1986}. Previous works have demonstrated that dimensionality reduction can improve not only computational cost but also the effectiveness of the data representation~\cite{Law:2019}~\cite{Su:2018}~\cite{Schwartz:2009}. In this context, Partial Least Squares (PLS) has presented remarkable results when compared to other dimensionality reduction methods~\cite{Schwartz:2009}. 
This is mainly due to the criterion through which PLS finds the low dimensional space, which is by capturing the relationship between independent and dependent variables. Another interesting aspect of PLS is that it can operate as a feature selection method, for instance, by employing Variable Importance in Projection (VIP)~\cite{Mehmood:2012}. The VIP technique employs score matrices yielded by NIPALS (the standard algorithm used for traditional PLS) to compute the importance of each feature based on its contribution to the generation of the latent space.
\begin{figure*}[!htb]
	\centering
	\subfigure[IPLS projection.] {\includegraphics[scale=0.4]{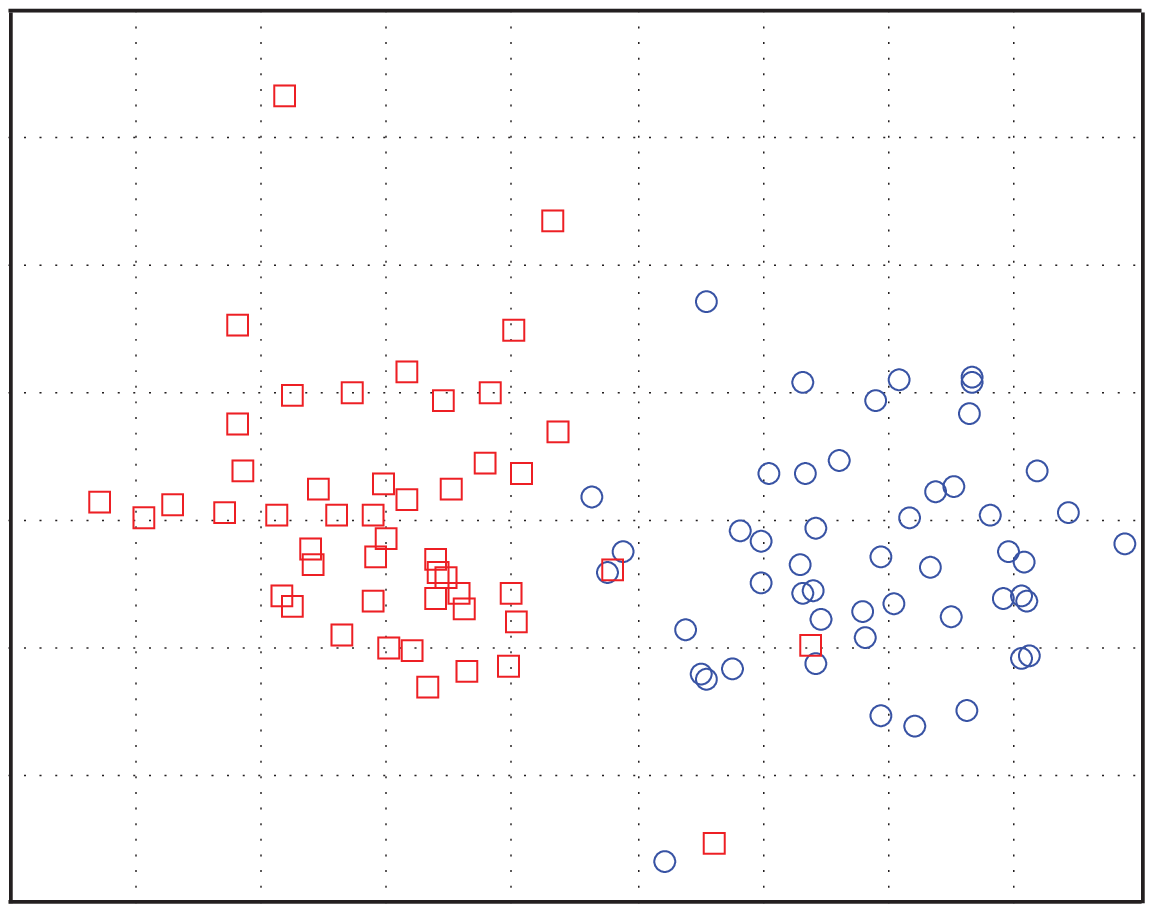}}
	\hspace{-33pt}
	\subfigure[SGDPLS projection.]
	{\includegraphics[scale=0.4]{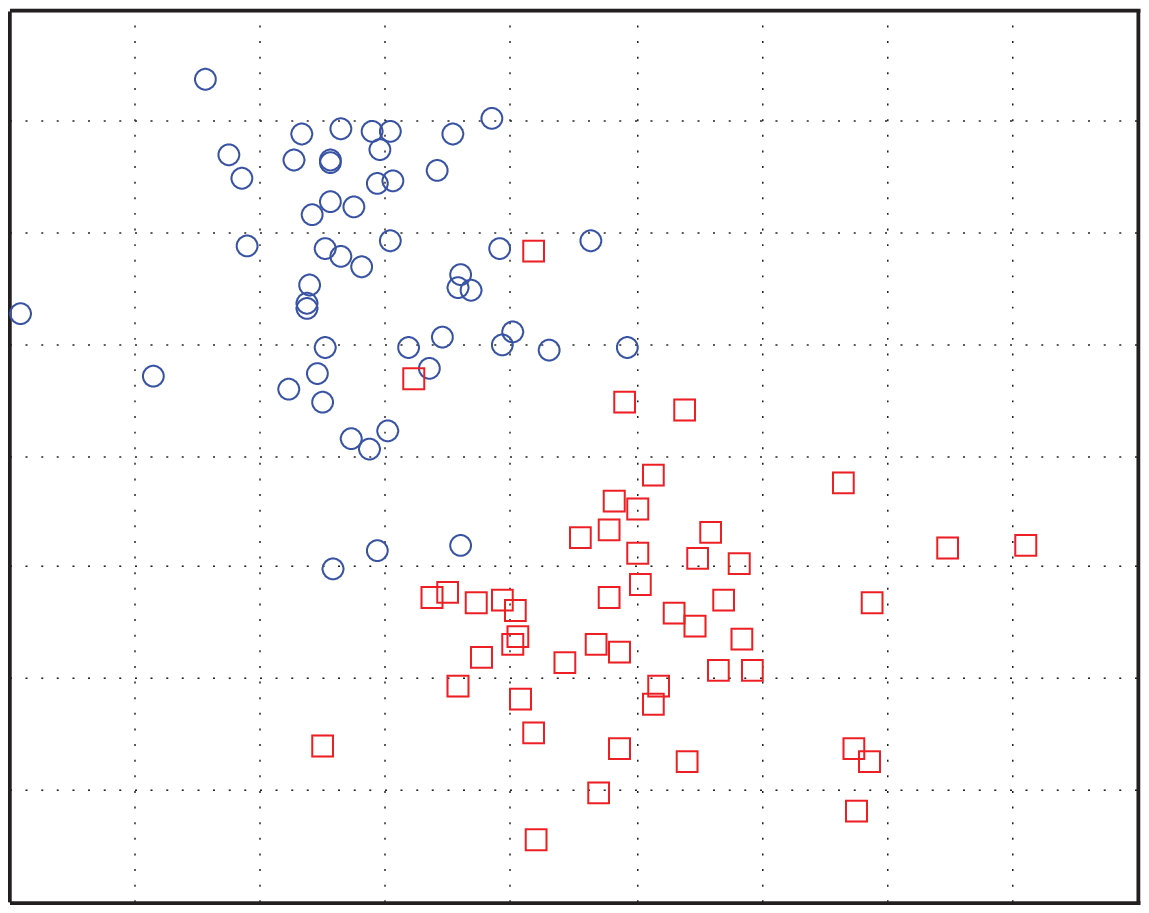}}
	\hspace{-33pt}
	\subfigure[CIPLS (Ours) projection.] {\includegraphics[scale=0.4]{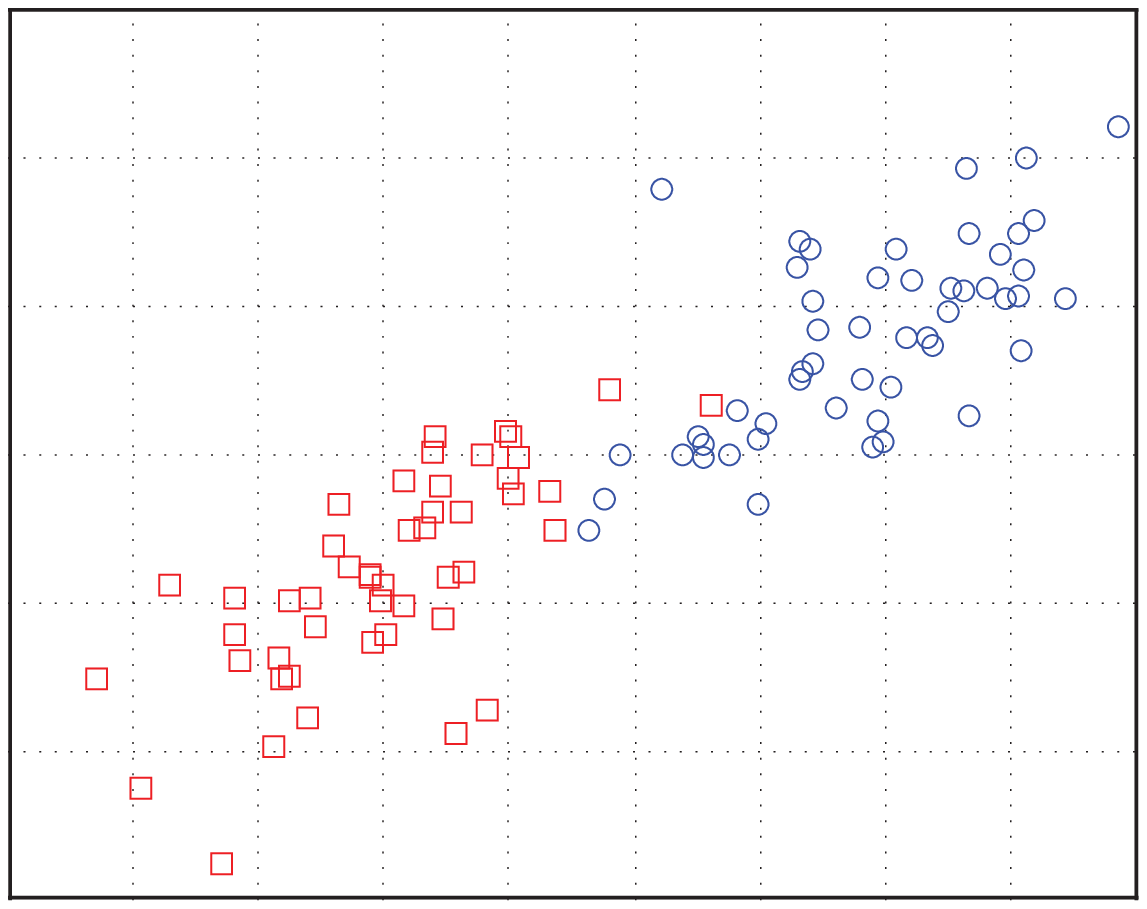}}
	\caption{Projection on the first (x-axis) and second (y-axis) components using different dimensionality reduction techniques. Our method (CIPLS) separates the feature space better than IPLS and SGDPLS, which are state-of-the-art incremental PLS-based methods. For IPLS and SGDPLS, the class separability is effective only on a single dimension of the latent space, while for CIPLS it is retained on both dimensions. Blue and red points denote positive and negative samples, respectively.}
	\label{fig:latent_space}
\end{figure*}

Despite achieving notable results, PLS is not suitable for large datasets, such as  ImageNet~\cite{imagenet}, since it requires all the data to be in memory in advance, which is often impractical due to hardware limitations. Additionally, this requirement prevents us from employing PLS on streaming applications, where the data are being generated continuously. Such limitation is not particular to PLS, many dimensionality reduction methods, such as Principal Component Analysis (PCA) and Linear Discriminant Analysis (LDA), also suffer from this problem~\cite{Weng:2003, Alakkar:2019, Yang:2019}.

To handle the aforementioned problem, many works have proposed incremental versions of traditional dimensionality reduction methods. The idea behind these methods is to estimate the projection matrix using a single data sample (or a subset) at a time while keeping some properties of the traditional dimensionality reduction methods. A well-known class of incremental methods is the one based on Stochastic Gradient Descent (SGD)~\cite{Arora:2016}~\cite{Alakkar:2019}. These methods interpret dimensionality reduction as a stochastic optimization problem of an unknown distribution. As shown by Weng et al.~\cite{Weng:2003}, incremental methods based on SGD are computationally expensive, present convergence problems and require many parameters that depend on the nature of the data. To address this problem, Zeng et al.~\cite{Zeng:2014} proposed an efficient and low-cost incremental PLS (IPLS). In their work, the first dimension (component) of the latent space is found incrementally, while the other dimensions are estimated by projecting the first component onto the reconstructed covariance matrix, which is employed to address the issue of impractical memory requirements of a full covariance matrix.

Even though IPLS achieves better performance than SGD-based and other state-of-the-art incremental methods, the discriminability of its higher-order components (i.e., all except the first) is not preserved, as shown in Figure~\ref{fig:latent_space} (a), where it can be seen that the effectiveness of class separability of IPLS is restricted to the first dimension of the latent space. This behavior occurs because the higher-order components are estimated using only the independent variables, that is, they are based on an approximation of the covariance matrix $X^{\top}X$ (similar to PCA) instead of $X^{\top}Y$ employed in PLS. This can degrade the discriminability of the latent model since preserving the relationship between independent and dependent variables is an important property of the original  PLS~\cite{Geladi:1986}. It is important to emphasize that, for high-dimensional data, employing several components often provides better results~\cite{Schwartz:2009,Guo:2011,Guo:2013}, hence, IPLS might not be suitable for these cases.

Motivated by limitations and drawbacks in incremental PLS-based approaches, we propose a novel incremental method\footnote{https://github.com/arturjordao/IncrementalDimensionalityReduction}. Our method is based on the hypothesis that the estimation of higher-order components using the covariance matrix, as proposed by Zeng et al.~\cite{Zeng:2014}, is inadequate since the relationship between independent and dependent variables is lost. Therefore, to preserve this characteristic, we extend NIPALS~\cite{Abdi:2010} to avoid the computation of $X^{\top}Y$ and, consequently, enable it for incremental operation. Since our proposed extension is based on a simple algebraic decomposition, we preserve the simplicity and efficiency that makes NIPALS attractive, and we ensure that the relationship between independent and dependent variables is propagated to all components, differently from other methods.

As shown in Figure~\ref{fig:latent_space}, our method is capable of separating data classes better than IPLS, mainly on the second component (i.e., y-axis). Since the proposed method does not use the covariance matrix ($X^{\top}X$) to estimate higher-order components, we refer to it as \emph{Covariance-free Incremental Partial Least Squares} (CIPLS). Besides providing superior performance, our method can easily be extended as a feature selection technique since it provides all the requirements to perform VIP. Existing incremental PLS methods, on the other hand, require more complex techniques to operate as feature selection~\cite{Mehmood:2012}. 

We compare the proposed method on the tasks of face verification and image classification, where it outperforms several other incremental methods in terms of accuracy and efficiency. In addition, in the context of feature selection, we evaluate and compare the proposed method to state-of-the-art methods, where it achieves competitive results.
\section{Related Work}\label{sec:relatedwork}
%Computer Vision problems usually suffer from very high dimensional data. To overcome this problem we can apply two approaches: dimensionality reduction and feature selection. While the first combines the input to achieve a new data representation, the second select a subset of potential features from the original data.
%
To enable PCA to operate in an incremental scheme, Weng et al.~\cite{Weng:2003} proposed to compute the principal components without estimating the covariance matrix, which is {unknown and impossible} to be calculated in incremental methods. For this purpose, their method, named CCIPCA, updates the projection matrix for each sample $x$, replacing the unknown covariance matrix by the sample covariance matrix $xx^{\top}$. While CCIPCA provides a minimum reconstruction error of the data, it might not improve or even preserve the discriminability of the resulting subspace since label information is ignored (similarly to traditional PCA)~\cite{Martinez:2001}. 

To achieve discriminability, incremental methods based on Linear Discriminant Analysis (LDA) have been proposed~\cite{Hiraoka:2000}~\cite{Lu:2012}. In particular, this class of methods is less explored since they present issues such as the \emph{sample size problem}~\cite{Howland:2006}, which makes them infeasible for some tasks.
%https://scikit-learn.org/stable/modules/generated/sklearn.decomposition.LatentDirichletAllocation.html#sklearn.decomposition.LatentDirichletAllocation.partial_fit
%Falar porque nao usamos LDA-based ou realizar os experimentos com ILDA
%
Different from incremental LDA methods, incremental PLS methods are more flexible and present better results~\cite{Zeng:2014}. Motivated by this, Arora et al.~\cite{Arora:2016} proposed an incremental PLS based on stochastic optimization (SGDPLS), where the idea is to optimize an objective function using a single sample at a time. Similarly to Arora et al.~\cite{Arora:2016}, Stott et al.~\cite{Stott:2017} proposed applying stochastic gradient maximization on NIPALS, extending it for incremental processing. Even though they present promising results on synthetic data, their approach presented convergence problems when evaluated on real-world datasets. Thus, in this work, we consider only the approach by Arora et al.~\cite{Arora:2016}, which was the one that converged for several of the datasets evaluated and presented better results.

While SGDPLS is effective, as demonstrated by Weng et al.~\cite{Weng:2003} and Zeng et al.~\cite{Zeng:2014}, SGD-based methods applied to dimensionality reduction are computationally expensive and present convergence problems. In addition, this class of approaches requires careful parameter tuning and their results are often sensitive to the type of dataset~\cite{Weng:2003}.

To address convergence problems in SGD-based PLS, Zeng et al.~\cite{Zeng:2014} proposed to decompose the relationship between independent and dependent matrices (variables) into a sample relationship (i.e., a single sample with its label). This process is performed only to compute the first component, while the higher-order components are estimated by projecting the first component onto an approximated covariance matrix using a few PCA components. As we mentioned earlier, since traditional PCA cannot be employed in incremental methods, Zeng et al.~\cite{Zeng:2014} used CCIPCA to reconstruct the principal components of the covariance matrix. 

In contrast to existing incremental PLS methods, our method presents superior performance in both accuracy and execution time for estimation of the projection matrix, which is an important requirement for time-sensitive and resource-constrained tasks. In particular, considering the average accuracy across all tasks in our assessment, the proposed method outperforms IPLS and SGDPLS by $32.48$ and $24.83$ percentage points, respectively, when using only higher-order components. The reason for these results is the quality of our higher-order components, which keeps the discriminative properties of traditional PLS.
%\todo{precisa falar em qual dataset que teve essa melhora de performance}

Another line of research widely employed to reduce computational cost is feature selection. One of the most successful feature selection methods is the work by Roffo et al.~\cite{Roffo:2015}, which proposed to interpret feature selection as a graph problem. In their method, named \emph{infinity feature selection} (infFS), each feature represents a node in an undirected fully-connected graph and the paths in this graph represent the combinations of features. Following this model, the goal is to find the best path taking into account all the possible paths (in this sense, all the subsets of features) on the graph, by exploring the convergence property of the geometric power series of a matrix.
Improving upon this model, Roffo et al.~\cite{Roffo:2017} suggested quantizing the raw features into a small set of tokens before computing infFS. By using this pre-processing, their method (referred to as \emph{infinity latent feature selection} --- ilFS) achieved even better results than infFS. 
%\changed{Recently, Roffo et al.~\cite{Roffo:2020} presented computational cost-friendly version of infFS, which consider supervised (infFS$_\text{S}$) and unsupervised (infFS$_\text{U}$) scenarios.}
Recently, Roffo et al.~\cite{Roffo:2020} presented a more efficient version of infFS, which considers supervised (infFS$_\text{S}$) and unsupervised (infFS$_\text{U}$) scenarios.
Although the framework by Roffo et al.~\cite{Roffo:2015, Roffo:2017, Roffo:2020} achieved state-of-the-art results, in the context of neural network optimization, Jordao et al.~\cite{Jordao:2020} showed that PLS+VIP attains superior performance. We show that CIPLS+VIP achieves comparable results when compared to PLS+VIP and other state-of-the-art feature selection techniques.
\section{Proposed Approach}\label{sec:proposed_method}
In this section, we start by describing the traditional Partial Least Squares (PLS). Then, we present the proposed \emph{Covariance-free Incremental Partial Least Squares} (CIPLS) and the Variable Importance in Projection (VIP) technique, which enables PLS and CIPLS to be employed for feature selection. Unless stated otherwise, let $X \in \mathbb{R}^{n\times m}$ be the matrix of independent variables denoting $n$ training samples in a $m$-dimensional space. Furthermore, let $Y \in \mathbb{R}^{n\times 1}$ be the matrix of dependent variables representing the binary class label. Finally, let $x_n \in \mathbb{R}^{1\times m}$ and $y_n \in \mathbb{R}^{1 \times 1}$ be a single sample of $X$ and $Y$, respectively. We highlight that, in the context of streaming data, $x_n$ is a data sample acquired at time $n$.

%\vspace{1mm}
\subsection{Partial Least Squares}
Given a high $m$-dimensional space, PLS finds a projection matrix $W (w_1, w_2, ..., w_c)$, which projects this space onto a low $c$-dimensional space, where $c \ll m$. For this purpose, PLS aims at maximizing the covariance between the independent and dependent variables such that, besides reducing dimensionality, it preserves the discriminability of the data, which is essential for classification tasks. Formally, PLS constructs $W$ such that
\begin{equation}\label{eq:pls}
w_i = maximize(\mathrm{cov}(Xw, Y)), \text{s.t} \lVert w \rVert = 1,
\end{equation}
where $w_i$ denotes the $i$th component of the $c$-dimensional space. The exact solution to Equation~\ref{eq:pls} is given by 
\begin{equation}\label{eq:pls2}
w_i = \frac{X^{\top}Y}{\lVert X^{\top}Y \rVert}.
\end{equation}

From Equation~\ref{eq:pls2}, we can compute all the $c$ components using either  Nonlinear Iterative Partial Least Squares (NIPALS)~\cite{Abdi:2010} or Singular Value Decomposition (SVD). Most works employ NIPALS since it is capable of finding only the $c$ first components, while SVD always finds all the $m$ components, being computationally prohibitive for large datasets~\cite{Xu:2019, Maalouf:2019}.
%inefficient compared to NIPALS~\cite{Abdi:2010}.

\subsection{Covariance-free Incremental Partial Least Squares}
The core idea in our method is to ensure that, as in traditional PLS, the relationship between independent and dependent variables (Equation~\ref{eq:pls2}) is kept on all the $c$ components. To achieve this goal, our method works as follows. First, we need to center the data to the mean of the training samples $X$. However, different from traditional methods, in incremental approaches the mean is unknown since we cannot assume that all the data are known a priori~\cite{Weng:2003}~\cite{Zeng:2014}. To face this problem, we center the current data sample using an approximate centralization process~\cite{Weng:2003} which consists of estimating an incremental mean using the $n$th sample. According to Weng et al.~\cite{Weng:2003}, we can compute the incremental mean $\mu_n$ w.r.t. the $n$th data sample as
\begin{equation}\label{eq:z_score}
\mu_n = \frac{n-1}{n}\mu_{(n-1)} + \frac{1}{n}x_n.
\end{equation}

Once we have centralized the sample, the next step in our method is to compute the component $w_i$ following Equation~\ref{eq:pls2}. As we mentioned, $X$ and its respective $Y$ are unknown or are not in memory in advance, which prohibits us to apply Equation~\ref{eq:pls2} directly. However, as suggested by Zeng et al.~\cite{Zeng:2014}, we employ the following decomposition:
\begin{equation}\label{eq:incremental_pls}
X^TY = \sum_{k=1}^{n-1}x^T_ky_k + x^T_ny_n.
\end{equation}
By replacing $X^{\top}Y$ in Equation~\ref{eq:pls2} by Equation~\ref{eq:incremental_pls}, it is possible to calculate the $i$th component of PLS considering a single sample at a time. In other words, Equation~\ref{eq:incremental_pls} enables to compute $w_i$ incrementally.

To compute the higher-order components {($w_i$, \mbox{$i > 1$)}}, we employ a \emph{deflation} process that consists of subtracting the contribution of the current component on the sample before estimating the next component. Following the NIPALS algorithm, the deflation process works as follows
\begin{equation}\label{eq:deflation1}
t = Xw_{i},
\end{equation}
\begin{equation}\label{eq:deflation2}
p = X^{\top}t, q=Y^{\top}t,
\end{equation}
\begin{equation}\label{eq:deflation3}
X = X-tp^{\top}, Y = Y- tq^{\top},
\end{equation}
where $t$ denotes the projected samples onto the current component $w_i$, and $p$ and $q$ represent the loadings of this projection. It should be noted that while $t$ works in an incremental scheme (since we can project one sample at a time), $p$ and $q$ cannot be computed since $X$ and $Y$ are neither known nor are in memory in advance. However, in light of Equation~\ref{eq:incremental_pls}, we can decompose $p$ and $q$ as
\begin{equation}\label{eq:incremental_scores}
\begin{split}
p = \sum_{k=1}^{n-1}x_k^{\top}t_k + x_n^{\top}t_n,
q= \sum_{k=1}^{n-1}y_k^{\top}t_k + y_n^{\top}t_n.
\end{split}
\end{equation}
By embedding Equation~\ref{eq:incremental_scores} in the deflation process, we can remove the contribution of the current component and repeat the process to compute a single component $w_i$ (as we argued before).
%
%Observe that Equation~\ref{eq:deflation3} can be computed sample-by-sample working, therefore, in an incremental scheme. 
Observe that Equation~\ref{eq:deflation3} deflates each sample by its reconstructed value. This way, Equation~\ref{eq:deflation3} can be computed sample-by-sample, working in an incremental scheme. With this formulation, we are now capable of computing the $c$ components incrementally. Algorithm~\ref{alg::cipls} summarizes the steps of our method. It should be mentioned that the matrices $W$, $P$ and $Q$ are initialized with zeros.

According to Algorithm~\ref{alg::cipls}, the proposed method maintains the property of capturing the relationship between $X$ and $Y$ for all components (step $4$ in Algorithm~\ref{alg::cipls}). In addition, since we compute all components at once, our method has a time complexity of $O(ncm)$, where $n$, $c$ and $m$ denote the number of samples, number of components, and dimensionality of the data, respectively.

\begin{algorithm}[!htb]
\SetAlgoLined
\DontPrintSemicolon
\caption{CIPLS Algorithm.}
%http://www.jmlr.org/papers/volume2/rosipal01a/rosipal01a.pdf for PLS2
\label{alg::cipls}
	\SetKwInOut{Input}{Input}
	\SetKwInOut{Output}{Output}
	\Input{$n$th data sample $x_n$ and its label $y_n$\\Number of components $c$\\
	Projection matrix $W_{(n-1)} \in \mathbb{R}^{m \times c}$\\Loading matrix $P_{(n-1)}$ $\in \mathbb{R}^{m \times c}$\\ Loading matrix $Q_{(n-1)}$ $\in \mathbb{R}^{1 \times c}$}
	\Output{Updated matrices $W$, $P$ and $Q$}
    \BlankLine
%    \If{$W, P$ and $Q$ were not initialized}
%    {
%    	\BlankLine
%    	Initialize $W, P$ and $Q$ with zeros
%    }
    \BlankLine
    Update $\mu_n$ using Equation~\ref{eq:z_score}
    \BlankLine
    $\bar{x}_n = x_n- \mu_n$
    \BlankLine
    \For{ $i=1$ \textbf{to} $c$}{
    $w_{i} = \bar{x}^{\top}_ny_n + w_{i(n-1)}$, where $w_i \in W$
    \BlankLine
    $t_n = \frac{\bar{x}_nw_i}{\lVert \bar{x}_nw_i \rVert}$
       	\BlankLine
    $p_i = \bar{x}^{\top}_nt_n + p_{i(n-1)}$, where $p_i \in P$
    \BlankLine
    $q_i = y_n^{\top}t_n + q_{i(n-1)}$, where $q_i \in Q$
    \BlankLine
    $\bar{x}_n = \bar{x}_n - t_np_i^{\top}$
    \BlankLine
    $y_n = y_n - t_nq_i^{\top}$
    }
\end{algorithm}

%\vspace{1mm}
\subsection{CIPLS for Feature Selection}
An advantage of PLS is that, after estimating the projection matrix $W$, it is possible to estimate the importance of each feature, enabling PLS to operate as a feature selection method. For this purpose, it is possible to employ Variable Importance in Projection (VIP), which estimates the importance of each feature $f_j$ based on its contribution to yield the low dimensional space.
%\todo{nao entendi o que quer dizer com w.r.t. its contribution..., nao conecta com o que vem antes na sentenca}. 
According to Mehmood et al.~\cite{Mehmood:2012}, VIP is defined as
\begin{equation}\label{eq:vip}
%f_m = \sqrt{m \sum_{i=1}^{c}\frac{\left[q_{i}^2t_i^Tt_i\frac{W_{im}}{\lVert W_{i} \rVert ^2}\right]}{\sum_{i=1}^{c}q_{i}^2t_i^Tt_i}}.
%
f_j = \sqrt{m\sum_{i=1}^c q_{i}^2t_i^{\top}t_i(w_{ij}/\Vert w_i \Vert ^2)/\sum_{i=1}^c q_{i}^2t_i^{\top}t_i}.
\end{equation}

Once we have estimated the score of each feature, we can remove a percentage of features based on their scores. As can be verified in Algorithm~\ref{alg::cipls}, CIPLS preserves the ability of traditional PLS to be employed as a feature selection method via VIP (Equation~\ref{eq:vip}). It is important to emphasize that IPLS and SGDPLS cannot be used to compute VIP as they do not provide the loading matrix $Q$ ($q_1, q_2,..., q_c$). 
\begin{table*}[!!htb]
	\centering
	\renewcommand{\arraystretch}{1.2}
	\caption{Comparison of existing incremental methods in terms of accuracy. The symbol '--' denotes that it was not possible to execute the method on the respective dataset due to memory constraints or convergence problems (see the text). PLS denotes the use of the traditional PLS. The closer to the accuracy of the baseline (PLS), the better. The numbers enclosed in square brackets denote confidence interval ($95\%$ confidence).}
	\label{tab:main_results}
	\begin{tabular}{ccccc}
		\hline
		& LFW                          & YTF                          & \begin{tabular}[c]{@{}c@{}}ImageNet\\ 32$\times$32\end{tabular} & \begin{tabular}[c]{@{}c@{}}ImageNet\\ 224$\times$224\end{tabular} \\ \hline
		\multicolumn{1}{c|}{CCIPCA~\cite{Weng:2003}} & \multicolumn{1}{c|}{$89.87~[89.17, 90.55]$} & \multicolumn{1}{c|}{$81.48~[80.07, 82.88]$} & \multicolumn{1}{c|}{$40.30$}                             & $52.58$                                                    \\
		\multicolumn{1}{c|}{SGDPLS~\cite{Arora:2016}} & \multicolumn{1}{c|}{$90.60~[89.95, 91.24]$} & \multicolumn{1}{c|}{$83.22~[82.07, 84.36]$} & \multicolumn{1}{c|}{--}                                   & --                                                          \\
		\multicolumn{1}{c|}{IPLS~\cite{Zeng:2014}}   & \multicolumn{1}{c|}{$90.30~[89.60, 90.99]$} & \multicolumn{1}{c|}{$82.22~[80.96, 83.47]$} & \multicolumn{1}{c|}{$43.24$}                             & $65.74$                                                    \\
		\multicolumn{1}{c|}{\textbf{CIPLS (Ours)}}    & \multicolumn{1}{c|}{$91.78~[91.08, 92.47]$} & \multicolumn{1}{c|}{$84.10~[82.82, 85.37]$} & \multicolumn{1}{c|}{$43.31$}                                   & $67.09$                                                           \\ \hline
		\multicolumn{1}{c|}{PLS}  & \multicolumn{1}{c|}{$92.47~[91.87,93.05]$} & \multicolumn{1}{c|}{$85.96~[84.47, 87.44]$} & \multicolumn{1}{c|}{--}                             & --                                                   \\ \hline
	\end{tabular}
\end{table*}
\section{Experimental Results}\label{sec:experiments}
In this section, we first introduce the experimental setup and the tasks employed to validate the proposed method. Then, we present the procedure conducted to calibrate the parameters of the methods. Next, we compare the proposed method with other incremental partial least squares methods as well as with the traditional PLS. Afterwards, we present the influence of higher-order components on the classification performance. Finally, we discuss the time complexity of the methods, their performance on a streaming scenario and compare our method on the feature selection context.
%
%\changed{It is worth mentioning that our focus is on the behavior of the approaches, but not on pushing the state-of-the-art results.}

\vspace{1mm}
\noindent\textbf{Experimental Setup.} Following previous works~\cite{Donahue:2014, Razavian:2014, Azizpour:2016, Kornblith:2019}, we employ linear classifiers when using features from convolutional networks. Specifically, we use a linear SVM as employed by our main baseline~\cite{Zeng:2014}.
%Throughout the experiments, we use a linear SVM for binary classification (face verification) because, according to Zeng et al.~\cite{Zeng:2014}, a linear SVM coupled with dimensionality reduction is able to achieve remarkable results while being computationally efficient. In addition, it has been shown that simple classifiers when feed by features from convolutional networks are able to achieve results comparable to more sophisticated classifiers~\cite{Donahue:2014}~\cite{Razavian:2014}. 
%
For multi-class problems (image classification), on the other hand, we prefer to use a multilayer perceptron since it handles the multi-class problem naturally, avoiding the need for employing a binary classifier (e.g., SVM) on a one-versus-rest fashion, which would be computationally expensive. All experiments and methods were executed on an Intel Core i5-8400, 2.4 GHz processor with 16 GB of RAM.

To assess the differences in efficacy and efficiency among the compared methods, throughout the experiments we perform statistical tests based on a paired t-test using $95\%$ confidence~\cite{Raj:1990}. We highlight that the statistical tests were conducted only for face verification due to the computational cost of retraining (i.e., fine-tuning) the convolutional network for image classification, which is considerably high since we employ large-scale datasets.

\vspace{1mm}
\noindent
\textbf{Face Verification.} Given a pair of face images, face verification determines whether this pair belongs to the same person. For this purpose, we use a three-stage pipeline~\cite{Ranjan:2018, Chen:2018} as follows. First, we extract a feature vector of each face using a deep learning model. In this work, we use the feature maps from the last convolutional layer of the VGG16 model, learned on the VGGFaces dataset~\cite{Parkhi:2015}, as feature vector. Then, we compute the distance between the two feature vectors employing the $\ell_1$-distance metric. Finally, we present the result of the distance metric to a classifier.

We conduct our evaluation on two face verification datasets, namely Labeled Faces in the Wild (LFW)~\cite{Huang:2012} and Youtube Faces (YTF)~\cite{Wolf:2011}.

\vspace{1mm}
\noindent
\textbf{Image Classification.} 
%Image classification consists of deciding to which one of a given set of categories an image belongs. To this end, as suggested by previous works~\cite{Donahue:2014, Razavian:2014, Azizpour:2016, Kornblith:2019}, we extract features from the samples and feed these features to a classifier, which determines the category to which each image belongs. In this work, we use the features maps from the last convolutional layer of the VGG16 model as features.
%
For this task, we use features maps from the last layer of a VGG16 network as features. Moreover, we consider two versions of the ImageNet dataset, with images of size $224 \times 224$ and $32 \times 32$ pixels. The former is used since it is the original version of the dataset, while the latter is used because it has been demonstrated to be more challenging than the original version~\cite{Rebuffi:2017, Loshchilov:2017, Milacski:2019}. It is worth mentioning that the single difference between these versions of ImageNet is the image size.

\vspace{1mm}
\noindent
\textbf{Number of Components.} 
One of the most important aspects of dimensionality reduction methods is the number of components $c$ of the resulting latent space. Therefore, to choose the best number of components for each method, we vary $c$ from $1$ to $10$ and select the value for which the method achieved the highest accuracy on the validation set ($10\%$ of the training set). Once the best $c$ is chosen, we use the training and validation set to learn the projection method and the classifier. We repeat this process for each dataset.

\vspace{1mm}
\noindent
\textbf{Comparison with Incremental Methods.} This experiment compares the proposed CIPLS with other incremental dimensionality reduction methods. Table~\ref{tab:main_results} summarizes the results and shows that, on LFW, our method outperformed SGDPLS and IPLS by $1.18$ and $1.48$ percentage points (p.p.), respectively. Similarly, on YTF, CIPLS outperformed SGDPLS and IPLS by $0.88$ and $1.88$ p.p..

Finally, on the ImageNet dataset, the difference in accuracy compared to IPLS was of $0.07$ and $1.35$ p.p., for the $32 \times 32$ and $224 \times 224$ versions, respectively. It is important to mention that we do not consider SGDPLS on these datasets due to convergence problems and high computational cost. Furthermore, due to memory constraints, it was not possible to run the traditional PLS on ImageNet.
%\begin{figure}
%	\centering
%	\subfigure[SGDPLS]{\includegraphics[width=0.45\columnwidth]{Figures/loss_landscape_SGDPLS}}
%	\subfigure[IPLS]{\includegraphics[width=0.45\columnwidth]{Figures/loss_landscape_IPLS}}
%	\subfigure[CIPLS (Ours)]{\includegraphics[width=0.45\columnwidth]{Figures/loss_landscape_CIPLS}}
%	\caption{}
%	\label{fig:losslandscapecipls}
%\end{figure}

\vspace{1mm}
\noindent
\textbf{Comparison with Partial Least Squares.} As suggested by Weng et al.~\cite{Weng:2003}, we compare the incremental methods with the traditional approach as baseline (in our case, traditional PLS).  
%\todo{essa ultima parte quando fala do valor da medida, ficou perdida aqui, melhor colocar no paragrafo que discute a tabela}
% Moreover, by comparing the methods with the traditional PLS serves as an indication of how well the incremental methods are. 
%
According to Table~\ref{tab:main_results}, besides providing better results than IPLS and SGDPLS, CIPLS achieved the closest results to traditional PLS. For instance, on LFW, the difference in accuracy between PLS and CIPLS was $0.69$ p.p. while on YTF it was $1.86$ p.p.. In contrast, the difference in accuracy between PLS and SGDPLS is higher --- $1.87$ p.p. on LFW and $2.74$ p.p. on YTF. In addition, the difference in accuracy between PLS and IPLS is among the highest, $2.17$ and $3.74$ p.p. for the LFW and YTF datasets, respectively. In particular, the results for PLS and CIPLS are statistically equivalent, while IPLS and SGDPLS present results statistically inferior compared to PLS.

It should be noted that the results of IPLS are closer to CCIPCA than PLS  since only the first component of IPLS maintains the relationship between independent and dependent variables. On the other hand, the proposed method preserves this relation along higher-order components, which provides better discriminability, as seen in our results.
\begin{table}[!t]
	\centering
	\renewcommand{\arraystretch}{1.2}
	\caption{Accuracy of existing incremental methods when using only higher-order components. Values computed considering the average accuracy across all tasks in our assessment.}
	\label{tab:higher_order}
	\begin{tabular}{cc}
		\hline
		& Average Accuracy \\ \hline
		\multicolumn{1}{c|}{CCIPCA~\cite{Weng:2003}} & $63.48$ \\
		\multicolumn{1}{c|}{SGDPLS~\cite{Arora:2016}} & $58.41$ \\
		\multicolumn{1}{c|}{IPLS~\cite{Zeng:2014}} & $50.76$ \\ \hline
		\multicolumn{1}{c|}{CIPLS (Ours)} & $83.24$ \\ \hline
	\end{tabular}
\end{table}

\vspace{1mm}
\noindent
\textbf{Higher-order Components.} This experiment assesses the discriminability of the higher-order components of CIPLS compared to each of the other incremental methods. For this purpose, we follow a process suggested by Martinez~\cite{Martinez:2001}, which consists of removing the first component of the latent space before presenting the projected data to the classifier. This evaluates the performance of the remaining components, not only the first one which tends to be better. 

Table~\ref{tab:higher_order} shows the results. According to Table~\ref{tab:higher_order}, the proposed method outperforms IPLS by $32.48$ p.p.. Observe that when all the components are used, CIPLS outperforms IPLS by $1.17$ p.p.. This larger difference when removing the first component is an effect of the better discriminability achieved by the components extracted by CIPLS. As we have argued, CIPLS preserves the relationship between dependent and independent variables across higher-order components, yielding more accurate results. Compared to SGDPLS, CIPLS outperforms it by $24.83$ p.p. when using only the higher-order components.
\begin{table}[!t]
	\centering
	\renewcommand{\arraystretch}{1.2}
	\caption{Comparison of incremental dimensionality reduction methods in terms of time complexity  for estimating the projection matrix. $m$, $n$ denote dimensionality of the original data and number of samples, while $c$, $L$ and $T$ denote number of PLS components, number of PCA components and convergence steps.}
	\label{tab:time_complexity}
	\begin{tabular}{cc}
		\hline
		& Time Complexity \\ \hline
		\multicolumn{1}{c|}{CCIPCA~\cite{Weng:2003}} & $O(nLm)$ \\
		\multicolumn{1}{c|}{SGDPLS~\cite{Arora:2016}} & $O(Tcm)$ \\
		\multicolumn{1}{c|}{IPLS~\cite{Zeng:2014}} & $O(nLm+c^2m$) \\ \hline
		\multicolumn{1}{c|}{CIPLS (Ours)} & $O(ncm)$ \\ \hline
	\end{tabular}
\end{table}

\vspace{1mm}
\noindent
\textbf{Time Issues.} To demonstrate the efficiency of CIPLS, in this experiment, we compare its time complexity to compute the projection matrix with the incremental methods evaluated. Following Weng et al.~\cite{Weng:2003} and Zeng et al.~\cite{Zeng:2014}, we report this complexity w.r.t. dimensionality of the original data $m$, number of samples $n$, number of components $c$ and number of PCA components $L$ (required only by IPLS and CCIPCA). Table~\ref{tab:time_complexity} shows the time complexity of the methods.
\begin{figure}[!b]
	\centering
	\includegraphics[width=0.85\columnwidth]{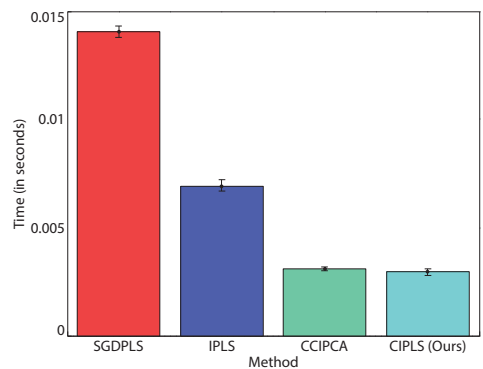}
	\caption{Average prediction time (in seconds) for estimating the projection matrix, lower values are better. Black bars denote the confidence interval.}
	\label{fig:time_issues}
\end{figure}
\begin{figure*}[!!htb]
	\centering
	\subfigure[Labeled Faces in the	Wild (LFW).] {\includegraphics[width=0.9\columnwidth]{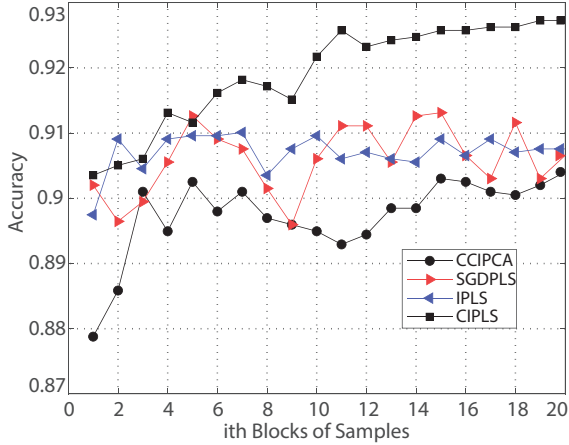}}
	\hspace{15pt}
	\subfigure[Youtube Faces (YTF).]
	{\includegraphics[width=0.9\columnwidth]{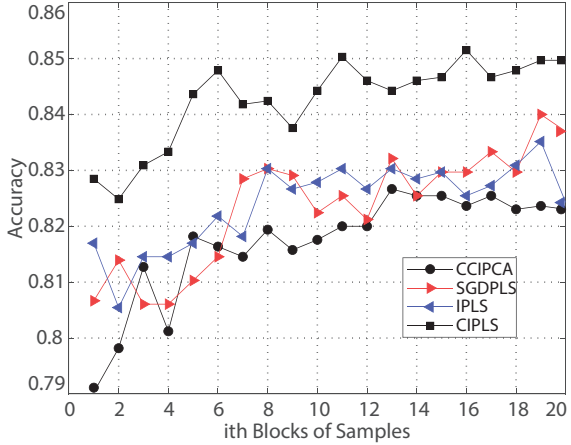}}
	\caption{Comparison of incremental methods on a streaming scenario. The x-axis denotes the data arriving sequentially.}
	\label{fig:streaming_application}
\end{figure*}

According to Table~\ref{tab:time_complexity}, CIPLS presents a low time complexity for estimating the projection matrix. The complexity of CIPLS is not only on the same class as CCIPCA, which is the fastest among the compared methods, but it also has a very small constant factor. This constant factor is the number of components, $c$ for CIPLS and $L$ for CCIPCA. Experimentally, we found that the optimal constant factor for PLS is negligible, $c=2$ resulted in the highest accuracies. While, for fairness, the same number of components was adopted for all methods in Table~\ref{tab:time_complexity}, typically $c<L$ on practical applications. This is a known advantage of PLS -- it has been shown to require substantially less components to achieve its optimal accuracy than PCA~\cite{Schwartz:2009}.

We also report the average computation time (considering $30$ executions) of the methods for estimating the projection matrix for one new sample. To make a fair comparison, we set $c=4$ for all methods and for the other parameters we use the values where the methods achieved the best results in validation. According to Figure~\ref{fig:time_issues}, SGDPLS is the slowest incremental PLS method, which is a consequence of its strategy for estimating the projection matrix, where for each sample the convergence step is run $T$ times. Our experiments showed that $T\geq100$ is required for good results. The computation time for estimating the projection matrix of our method was statistically equivalent (according to a paired t-test) to that of CCIPCA, which is the fastest among the incremental dimensionality reduction methods assessed. Moreover, CIPLS was statistically faster than IPLS and SGDPLS, demonstrating that it is the fastest among the compared incremental PLS methods.

\vspace{1mm}
\noindent
\textbf{Incremental Methods on the Streaming Scenario.} As we argued earlier, incremental methods can be employed on streaming applications, where the training data are continuously generated. To demonstrate the robustness of our method on these scenarios, we evaluate the methods on a synthetic streaming context, as proposed by Zeng et al.~\cite{Zeng:2014}. The procedure works as follows. First, the training data is divided into $k$ blocks, where $k=20$. The idea behind this process is to interpret each block as a new instance of arriving data. Then, we create a new training set and insert each $k$th block at a time. Each time we insert a new block, we learn the projection method and evaluate its accuracy on the testing set. For instance, when adding the tenth block, all the $1, 2,..., 10$ blocks are being used as training. It is important to mention that a block contains more than one sample, however, this does not modify the strategy of the incremental methods, which is to estimate the projection matrix by using a single sample at a time. 
%We execute the above-mentioned process on the LFW and YTF datasets. In particular, as observed by X, this process is not suitable on the cross validation, therefore, we change the evaluation protocol to holdout.

Figure~\ref{fig:streaming_application}~(a)~and~(b) show the results on the LFW and YTF datasets, respectively. On the LFW dataset, until the fifth block, it is not possible to determine the best method since the accuracy presents high variance, however, from the sixth block onwards, CIPLS outperforms all other methods. On the YTF dataset, our method achieves the highest accuracy for all blocks. These results show that the proposed method is more adequate for streaming applications than existing incremental PLS methods. 
\begin{table*}[!htb]
	\centering
	\renewcommand{\arraystretch}{1.2}
	\caption{Comparison of feature selection methods using different percentages of kept features.}
	\label{tab:feature_selection}
	\begin{tabular}{cccccccccc}
		\hline
		& \multicolumn{4}{c}{LFW} &  & \multicolumn{4}{c}{YTF} \\
		& \multicolumn{4}{c}{Percentage of Kept Features} &  & \multicolumn{4}{c}{Percentage of Kept Features} \\
		& $10$ & $15$ & $20$ & $50$ &  & $10$ & $15$ & $20$ & $50$ \\ \hline
		\multicolumn{1}{c|}{infFS~\cite{Roffo:2015}} & $91.92$ & $91.58$ & $92.03$ & $92.23$ &  & $86.64$ & $86.68$ & $87.14$ & $87.30$ \\
		\multicolumn{1}{c|}{ilFS~\cite{Roffo:2017}} & $92.03$ & $91.67$ & $92.25$ & $92.23$ &  & $86.60$ & $86.94$ & $86.84$ & $87.54$ \\
		\multicolumn{1}{c|}{infFS$_\text{U}$~\cite{Roffo:2020}} & $92.08$ & $91.70$ & $92.30$ & $92.15$ &  & $86.36$ & $86.60$ & $87.14$ & $87.16$ \\
		\multicolumn{1}{c|}{infFS$_\text{S}$~\cite{Roffo:2020}} & $91.80$ & $91.62$ & $91.62$ & $92.33$ &  & $86.12$ & $86.50$ & $86.80$ & $87.22$ \\
		\multicolumn{1}{c|}{PLS+VIP} & $92.05$ & $91.67$ & $92.13$ & $92.38$ &  & $86.70$ & $86.82$ & $87.18$ & $87.68$ \\ \hline
		\multicolumn{1}{c|}{CIPLS (Ours)+VIP} & $91.63$ & $91.55$ & $91.80$ & $92.18$ &  & $86.48$ & $86.92$ & $87.02$ & $87.40$ \\ \hline
	\end{tabular}
\end{table*}

\vspace{1mm}
\noindent
\textbf{Comparison with Feature Selection Methods.} Our last experiment evaluates the performance of CIPLS as a feature selection method. Table~\ref{tab:feature_selection} shows the results for different percentages of kept features on the LFW and YTF datasets.

According to Table~\ref{tab:feature_selection}, CIPLS is on par with the state-of-the-art feature selection techniques. For example, on LFW the difference in accuracy, on average, from CIPLS to infFS and ilFS is of $0.15$ and $0.25$ p.p., respectively. Compared to infFS$_\text{S}$ and infFS$_\text{U}$, this difference is $0.05$ and $0.26$ p.p., in this order.
Interestingly, on YTF for some percentages of kept features (e.g., $15\%$ and $50\%$), CIPLS outperforms ilFS, infFS$_\text{S}$ and infFS$_\text{U}$. We highlight that these methods were designed specifically for feature selection. 

Finally, the difference, on average, between CIPLS and PLS is of $0.26$ and $0.14$ p.p. on the LFW and YTF datasets. Moreover, the largest accuracy difference between PLS and CIPLS is only $0.4$ p.p., on LFW with $10\%$ of features kept. This result reinforces that the proposed decompositions to extend the NIPALS and enable the employment of VIP are a good approximation of the original method.

Based on the results shown, it is possible to conclude that, besides dimensionality reduction, the proposed method achieves state-of-the-art results in the context of feature selection. 
\section{Conclusions}\label{sec:conclusions}
This work presented a novel incremental partial least squares method, named \emph{Covariance-free Incremental Partial Least Squares} (CIPLS). The method extends the NIPALS algorithm for incremental operation and enables computation of the projection matrix using one sample at a time while still presenting the main property of traditional PLS, namely preserving the relation between dependent and independent variables. Compared to existing incremental partial least squares methods, CIPLS achieves superior performance besides being computationally efficient. In addition, different from previous incremental partial least squares, CIPLS can easily to operate as a feature selection method. In this context, the proposed method is able to achieve comparable results to the state of the art.
\section*{Acknowledgments}
The authors would like to thank the National Council for Scientific and Technological Development -- CNPq (Grants~140082/2017-4, ~438629/2018-3 and~309953/2019-7) and the Minas Gerais Research Foundation -- FAPEMIG (Grants~APQ-00567-14 and~PPM-00540-17.

%{\small
%\bibliographystyle{ieee_fullname}
%\bibliography{egbib}
%}
\bibliographystyle{ieee}
%\balance
\bibliography{refs} 

\end{document}